\documentclass[11pt, journal, final, twocolumn, letterpaper]{IEEEtran}

\usepackage{cite} 
\usepackage{amsmath, amssymb, amsfonts, amsthm}
\usepackage{mathtools}
\usepackage{array}
\usepackage{graphicx}
\usepackage{caption}
\usepackage{subcaption}
\usepackage{threeparttable} 
\usepackage{booktabs}
\usepackage{multirow}
\usepackage{enumitem}
\usepackage{todonotes}
\usepackage{soul}
\usepackage[ruled,vlined,linesnumbered]{algorithm2e}
\usepackage{tabularx}   
\usepackage{float}
\usepackage{microtype}
\usepackage[hidelinks]{hyperref}

\graphicspath{{Figs/}}

\begin{document}

\title{Multi-Task LLM with LoRA Fine-Tuning for Automated Cancer Staging and Biomarker Extraction}

\author{Jiahao~Shao,
        Anam~Nawaz~Khan,
        Christopher~Brett,
        Tom~Berg,
        Xueping~Li,
        and~Bing~Yao%
\thanks{J. Shao, A. N. Khan, T. Berg, X. Li, and B. Yao are with the Department of Industrial \& Systems Engineering, The University of Tennessee, Knoxville, TN 37996, USA (e-mail: byao3@utk.edu).}%
\thanks{C. Brett is with The University of Tennessee Medical Center, Knoxville, TN 37920, USA.}%
}

\maketitle

\begin{abstract}
Pathology reports serve as the definitive record for breast cancer staging, yet their unstructured format impedes large-scale data curation. While Large Language Models (LLMs) offer semantic reasoning, their deployment is often limited by high computational costs and hallucination risks. This study introduces a parameter-efficient, multi-task framework for automating the extraction of Tumor-Node-Metastasis (TNM) staging, histologic grade, and biomarkers. We fine-tune a Llama-3-8B-Instruct encoder using Low-Rank Adaptation (LoRA) on a curated, expert-verified dataset of 10,677 reports. Unlike generative approaches, our architecture utilizes parallel classification heads to enforce consistent schema adherence. Experimental results demonstrate that the model achieves a Macro F1 score of 0.976, successfully resolving complex contextual ambiguities and heterogeneous reporting formats that challenge traditional extraction methods including rule-based natural language processing (NLP) pipelines, zero-shot LLMs, and single-task LLM baselines. The proposed adapter-efficient, multi-task architecture enables reliable, scalable pathology-derived cancer staging and biomarker profiling, with the potential to enhance clinical decision support and accelerate data-driven oncology research.
\end{abstract}

\begin{IEEEkeywords}
Large language models (LLMs), Clinical NLP, Breast cancer, Tumor staging, Multi-task learning.
\end{IEEEkeywords}

\section{Introduction}
    Breast cancer remains the most diagnosed malignancy globally, accounting for substantial morbidity and mortality in women worldwide \cite{siegel2024cancer}. Accurate disease characterization is the cornerstone of effective treatment planning and prognostic assessment \cite{amin2017ajcc}. This specifically entails determining the Tumor-Node-Metastasis (TNM) stage, which quantifies disease extent based on tumor size (T), lymph node involvement (N), and metastasis (M), and biomarker status, including estrogen receptor (ER), progesterone receptor (PR), and human epidermal growth factor receptor 2 (HER2). While this critical information is codified in pathology reports, these documents exist primarily as unstructured free-text narratives or scanned PDFs within electronic health records (EHRs). Consequently, cancer registries often rely on manual data abstraction, a process that is notoriously resource-intensive and prone to human error \cite{alawad2020automatic,wang2024rule}.

    To automate this curation process, Natural Language Processing (NLP) has been widely explored. This approach mirrors a broader shift in healthcare systems engineering toward automated data-driven frameworks that achieve high efficacy in predicting complex clinical conditions. However, early NLP efforts relied heavily on rule-based systems and keyword matching, which offer high precision for standardized formats but exhibit brittle performance when distinguishing complex linguistic phenomena such as negation (e.g., ``No evidence of invasive carcinoma'') or fractured syntax caused by Optical Character Recognition (OCR) errors \cite{wu2020deep, yim2016natural}. Subsequent transformer-based models \cite{wang2025transformer,wang2026muse}, such as Bidirectional Encoder Representations from Transformers (BERT) and its clinical variants (BioBERT, ClinicalBERT), significantly improved semantic extraction from clinical text by enabling contextualized language representations\cite{devlin2019bert, lee2020biobert, alsentzer2019publicly}. However, these encoder-only architectures remain limited in their ability to capture the broad linguistic variability and complex dependencies present in clinical narratives. In particular, BERT models rely on masked language modeling objectives and typically require extensive task-specific fine-tuning, while operating within relatively constrained context windows (e.g., 512 tokens) and modest model scales (e.g., 110 million parameters for BERT-base). These characteristics restrict their capacity to model long-range relationships and perform flexible reasoning across multi-document clinical records. 
    
    LLMs represent a paradigm shift in clinical NLP. Recent studies have demonstrated the transformative potential of models like GPT-4 and Llama-3 for zero-shot extraction of structured data \cite{thirunavukarasu2023large}. For instance, Lee et al. \cite{lee2025using} successfully deployed local LLMs for surgical pathology extraction, and Garcia-Carmona et al. \cite{garcia2025leveraging} utilized prompt engineering to parse heterogeneous reports. However, these generative approaches encounter significant hurdles in production environments. Treating extraction as an open-ended text generation task introduces latency and the risk of hallucinating invalid formats (e.g., generating \texttt{Stage 2} when the schema requires \texttt{pT2}) \cite{agrawal2022large}. Furthermore, prompt-based methods often fails to generalize across diverse hospital templates without resource-intensive full fine-tuning, which remains computationally prohibitive for standard clinical infrastructure \cite{ding2023parameter}. While recent work has explored fine-tuning for extraction \cite{liu2025human}, relying on generative decoding introduces unavoidable stochasticity, forcing the use of complex, fragile post-processing logic to map probabilistic text tokens back to rigid clinical variables.

    In this paper, we propose a parameter-efficient, multi-task framework that repurposes the open-weight Llama-3-8B-Instruct architecture as a discriminative encoder. First, we replace the causal language modeling head with parallel, task-specific classification layers. We utilize the final-layer hidden representation of the input sequence as a fixed-dimensional embedding, which is mapped to task-specific outputs via parallel classification heads. By replacing autoregressive decoding with direct discriminative prediction, this formulation removes sampling-induced variability and produces outputs consistent with predefined clinical schemas. Second, we employ Low-Rank Adaptation (LoRA) \cite{hu2022lora} to enable high-performance fine-tuning, significantly reducing computational overhead compared to full-parameter fine-tuning. Third, we integrate an Multi-Task Learning (MTL) strategy to facilitate inductive knowledge transfer across correlated clinical variables. This joint optimization effectively mitigates the impact of class imbalance and enables the model to capture complex inter-task dependencies inherent in oncology narratives. We evaluate our proposed framework on a large-scale dataset of pathology narratives. Experimental results demonstrate that our framework significantly outperforms rule-based NLP and alternative Parameter-Efficient Fine-Tuning (PEFT) strategies in the high-fidelity extraction of structured oncology variables.
    
\section{Research Background}

    \subsection{Evolution of Clinical NLP and Generative Paradigm}
        The extraction of structured variables from unstructured pathology reports has evolved through three distinct methodological generations. The first generation relied on symbolic NLP, utilizing rule-based systems, Regular Expressions (regex), and dictionary lookups (e.g., UMLS, SNOMED CT) to identify clinical entities \cite{savova2010mayo, bodenreider2004unified, donnelly2006snomed}. While these systems offer high precision for standardized reporting templates, they are brittle: minor deviations in phrasing, typos, or OCR artifacts often lead to significant recall degradation \cite{alawad2020automatic}.
        
        The second generation marked the shift toward statistical learning and early deep learning, specifically discriminative models based on the Transformer architecture \cite{vaswani2017attention}. Domain-specific pre-trained models such as BioBERT \cite{lee2020biobert} and ClinicalBERT \cite{alsentzer2019publicly} demonstrated superior performance by learning contextual word representations from large biomedical corpora. However, these encoder-only models often lack the reasoning depth required to resolve complex long-range dependencies in multi-page pathology narratives. For instance, correctly staging a tumor often requires synthesizing information from the \texttt{Gross Description} (tumor size) and \texttt{Final Diagnosis} (histology) sections, a reasoning task that frequently exceeds the capacity of smaller BERT-based architectures \cite{wu2020deep}.
        These limitations paved the way for the third generation: LLMs with billions of parameters, such as GPT-4 and Llama-3. These models possess emergent reasoning capabilities that enable zero-shot or few-shot information extraction \cite{agrawal2022large}. Recent work has attempted to harness this potential for oncology; for example, Wang et al. \cite{wang2024using} utilized prompt engineering to automate cancer registry coding, while Lee et al. \cite{lee2025using} benchmarked local LLMs for surgical pathology extraction.

        Despite their promise, current studies highlight critical gaps in the generative paradigm. Lee et al. \cite{lee2025using} noted that while local LLMs protect privacy, they often struggle with consistent schema adherence compared to larger cloud-based models. Similarly, Wang et al. \cite{wang2024using} observed that prompt-based extraction frequently yields unstructured or variable outputs, necessitating complex post-processing logic to map predictions back to standardized registry codes (e.g., normalizing heterogeneous staging notation). These findings underscore that the dominant generative paradigm, where extraction is framed as probabilistic text generation, introduces unavoidable stochasticity and hallucination risks \cite{thirunavukarasu2023large, wornow2023shaky}. Fundamentally, generative decoding optimizes for the most probable next token rather than the most factually accurate category, rendering it suboptimal for rigid clinical databases. These limitations suggest that while the reasoning depth of LLMs is necessary, the extraction task is better framed as a discriminative classification problem rather than a generative one. This approach allows the framework to be guided by clinical domain knowledge, mirroring the constrained modeling strategies used to ensure the reliability of high-dimensional medical models. 

    \subsection{Parameter-Efficient Fine-Tuning and Multi-Task Learning}
        To address the limitations of prompt-based extraction, fine-tuning aligns the internal representations of LLM with the specific target task. However, fully fine-tuning models with over 7 billion parameters requires massive computational resources, often exceeding the memory capacity of standard hospital workstations. PEFT offers a solution by updating only a small subset of parameters. Strategies such as Prefix Tuning \cite{li2021prefix} and $(IA)^3$ (Infused Adapter by Inhibiting and Amplifying Inner Activations) \cite{liu2022few} utilize trainable vectors to steer the frozen activation patterns toward task-specific subspaces. Prefix Tuning achieves this by prepending learnable virtual tokens to the input sequence. In contrast, $(IA)^3$ introduces learnable vectors to rescale internal activations within the attention and feed-forward blocks via element-wise multiplication. Because this influence is mediated through activation modulation rather than direct weight reconfiguration, the expressive capacity of $(IA)^3$ is potentially constrained when substantial internal adaptation is required for complex clinical domain shifts.
        
        To enable more flexible yet still efficient adaptation, LoRA \cite{hu2022lora} modifies the parameterization strategy itself. LoRA introduces trainable low-rank decomposition matrices directly into selected weight matrices of the transformer architecture. This approach allows task-specific updates to be applied at the weight level while maintaining a compact parameter footprint. By exploiting the empirical observation that fine-tuning updates often reside in a low-dimensional subspace, LoRA approximates full fine-tuning with orders-of-magnitude fewer trainable parameters, achieving a stronger balance between adaptation capacity and computational efficiency.
        
        However, a significant challenge arises when applying LoRA to clinical datasets with skewed class distributions, which is a common issue in healthcare data analytics \cite{wang2024multi, shao2025multi,xie2024automated,wang2023hierarchical,wang2021multi}. Recent studies identify a critical vulnerability in standard LoRA: a tendency to exacerbate ``Catastrophic Inheritance,'' where pre-existing biases and skewed data distributions are amplified through the low-rank bottleneck \cite{chang2024ba}. This phenomenon is often driven by representation collapse within the adapters, where the constrained parameter space prioritizes global majority patterns at the expense of task-specific nuances, leading to performance deterioration on tail (minority) classes \cite{shi2023parameter}. In oncology datasets, where certain tumor categories are inherently long-tailed, this constraint can cause the adapters to overfit to the majority distribution, degrading performance on rare but clinically significant cases.
        
        To address these distributional challenges, we propose to integrate LoRA adaptation with MTL. By employing a hard parameter-sharing architecture \cite{ruder2017overview}, the low-rank adapters are shared across all extraction targets. This shared representation forces the model to learn a unified, regularized semantic space, effectively counteracting majority-bias by increasing the ``effective sample size'' for shared features. This facilitates inductive knowledge transfer where data-rich tasks (e.g., ER/PR status) stabilize the feature extraction for tasks with severe class imbalance (e.g., HER2 status) \cite{zhang2021survey}. In the biomedical NLP domain, such multi-task approaches have demonstrated significant efficacy in outperforming single-task baselines \cite{crichton2017neural}. In breast cancer staging specifically, where variables are intrinsically correlated, such as advanced tumor extent (T Stage) typically correlating with the likelihood of lymph node involvement (N Stage), jointly learning these tasks allows the framework to resolve linguistic ambiguities in data-scarce classes by leveraging the collective semantic strength of the entire report context. 
    
\section{Research Methodology}    
    We define our dataset formally as $\mathcal{D}=\{(\boldsymbol{x}^{(p)}, \boldsymbol{y}^{(p)})\}_{p=1}^S$, where $\boldsymbol{x}^{(p)}$ represents the input sequence for patient $p$. The target space is a composite vector $\boldsymbol{y}^{(p)} \in \mathcal{Y}_T \times \mathcal{Y}_N \times \dots \times \mathcal{Y}_{HER2}$, corresponding to the joint distribution of seven clinical variables: T Stage (tumor size), N Stage (lymph node involvement), M Stage (distant metastasis), histologic grade (cell differentiation), and three biomarkers (ER, PR, and HER2). The learning objective is to approximate the function $f_\theta: \boldsymbol{x}^{(p)} \to \boldsymbol{y}^{(p)}$ that minimizes the prediction error across all task heads simultaneously.

\begin{figure*}[t]
        \centering
        \includegraphics[width=0.85\textwidth]{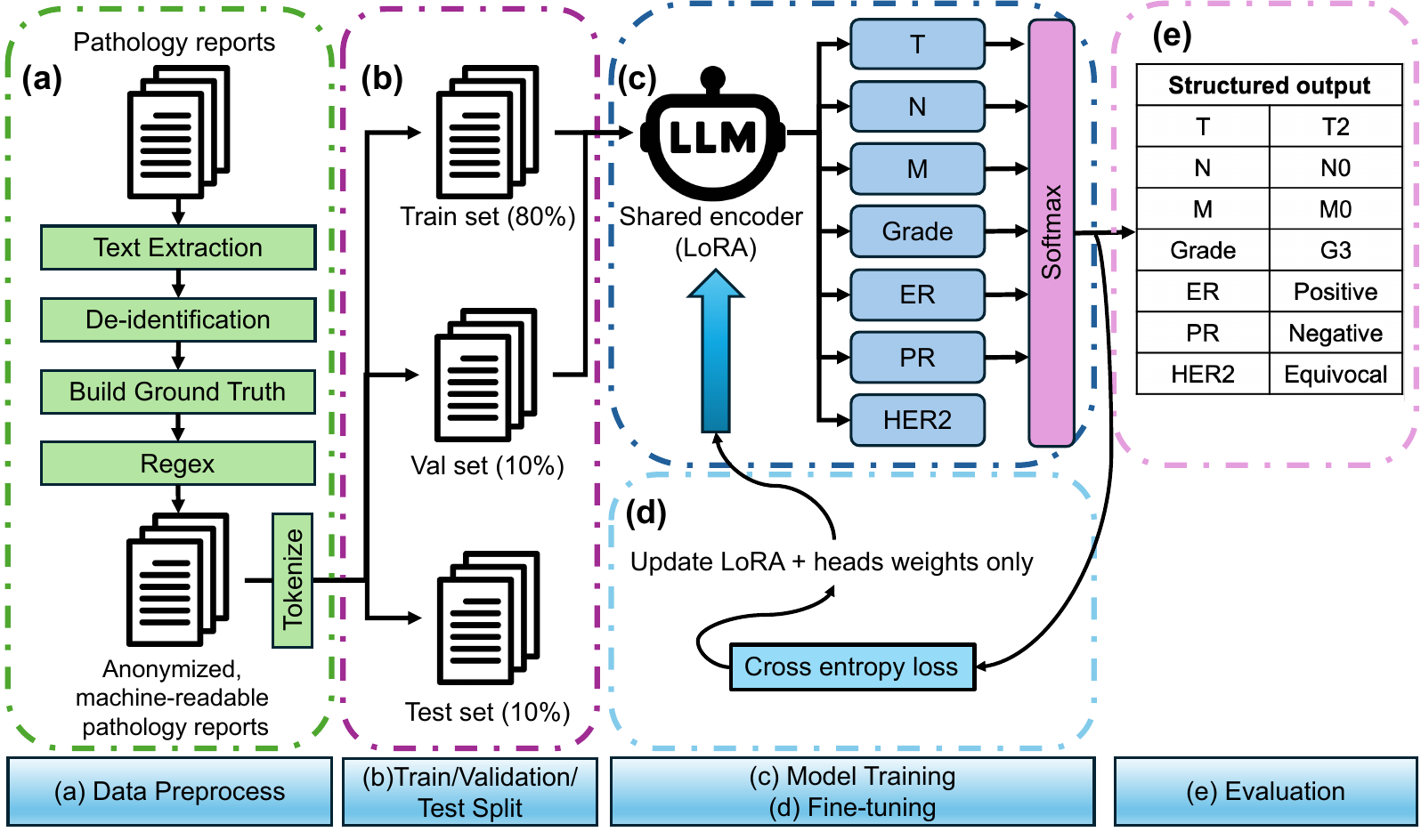} 
        \caption{
        Detailed architecture of the proposed Multi-Task LLM framework. (a) Data Preprocessing: Raw pathology reports undergo native digital text extraction, de-identification, and human-verified ground truth generation before tokenization. (b) Data Splitting: The dataset is partitioned into training (80\%), validation (10\%), and testing (10\%) sets to support model development and unbiased evaluation. (c) Model Architecture: A shared Llama-3 encoder (fine-tuned via LoRA) extracts high-dimensional semantic features from the terminal hidden state, which are simultaneously projected into parallel task-specific classification heads (T, N, M, Grade, ER, PR, HER2). (d) Optimization Loop: The training process minimizes a joint cross-entropy loss, updating only the low-rank adapters and head weights while keeping the heavy LLM backbone frozen. (e) Evaluation: The system produces a structured output vector containing the predicted class for each clinical variable, providing a comprehensive TNM and biomarker profile.
        }
        \label{fig:framework}
    \end{figure*}

    Fig.~\ref{fig:framework} shows our Multi-Task LLM framework, comprising five integrated phases. The pipeline begins with Data Preprocessing (a), combining text extraction, de-identification, and iterative regex bootstrapping to generate the ground-truth dataset. Data Splitting (b) employs stratified partitioning to preserve class distributions for robust evaluation under class imbalance. The Model Architecture (c) employs a shared Llama-3-8B-Instruct encoder to extract high-dimensional semantic features from the terminal hidden state, which are then projected into parallel task-specific classification heads. Next, the Optimization Loop (d) uses LoRA to minimize joint cross-entropy loss by updating only the low-rank adapter matrices while keeping the pre-trained backbone frozen. Finally, Evaluation (e) generates a structured prediction vector, consolidating the cancer staging and biomarker status into a unified clinical profile. Each of the components is described in detail in the following subsections.

    \subsection{Data Preprocessing and Ground Truth Construction}
        We digitize pathology documents via dual-path extraction, which are segmented into site-specific blocks using regex. Labels are generated through human-in-the-loop bootstrapping, incorporating an expert audit of 7,481 (70.1\%) reports. Full technical details are documented in Supplementary Appendix I.

        \subsubsection{Structured Input Representation}
            To optimize efficiency, we apply semantic window-slicing ($\pm 5$ lines) around prioritized clinical headers (e.g., \texttt{FINAL DIAGNOSIS}). Narratives are mapped to a four-block schema (\textit{Field, Classes, Report, Hints}) to bridge text and extraction heads. \textit{Hints} provide localized anchors via pattern matching, facilitating latent evidence synthesis without rigid constraints. The complete logic for the deterministic hint generation engine is detailed in Supplementary Appendix I.
            
            \begin{center}
            \fbox{%
            \begin{minipage}{\dimexpr\columnwidth-2\fboxsep-2\fboxrule\relax}
                \small\ttfamily
                \textbf{Field:} \{Task Name\} \\
                \textbf{Classes:} \{Fixed Label Set\} \\
                \textbf{Report:} \{Sliced Report Text\} \\
                \textbf{Hints:} \\
                \textit{\{Key:Value Regex Matches (e.g., ``T\_hint: pT1c'', ``HER2\_hint: IHC 2+, FISH amplified'')\}}
            \end{minipage}}
            \end{center}

    \subsection{Multi-Task LLM Architecture with LoRA}
    \label{sec:mtl_lora}
        The core of our predictive framework is a pre-trained LLM, i.e., the Llama-3-8B-Instruct architecture \cite{dubey2024llama}, which serves as a shared semantic encoder. Llama-3-8B-Instruct demonstrates state-of-the-art performance on reasoning and comprehension benchmarks among models with fewer than 10 billion parameters. Unlike standard base models optimized solely for token perplexity, the ``Instruct'' variant utilizes semantic representations refined through instruction fine-tuning and Reinforcement Learning from Human Feedback (RLHF) \cite{ouyang2022training}. Mathematically, this alignment process induces a more structured manifold in the latent space, where self-attention weights are conditioned to anchor on instruction-relevant tokens \cite{xiao2023efficient} (e.g., our structured \textit{Field} and \textit{Hints} headers). This creates a linearly separable foundation where high-level task intent is encoded directly into the hidden states \cite{saglam2025large}. For our application, this architecture facilitates the disambiguation of clinical concepts, such as distinguishing current T Stage from historical data, by leveraging a latent space that is pre-aligned to structured formatting. This provides a robust foundation for complex extraction than standard base models, which lack such intent-driven conditioning.
        
        Notably, our approach departs from standard auto-regressive text generation. Instead of retaining the original causal language modeling head, we remove it and repurpose the Llama-3 backbone as a high-capacity feature extractor. Specifically, we extract the contextualized hidden states from the final transformer layer and utilize the terminal hidden state of the input sequence to form a dense semantic representation of the input. This task-agnostic context vector is then supplied to parallel classification heads for structured multi-task prediction.

        \begin{figure}[t]
            \centering
            \includegraphics[width=\columnwidth]{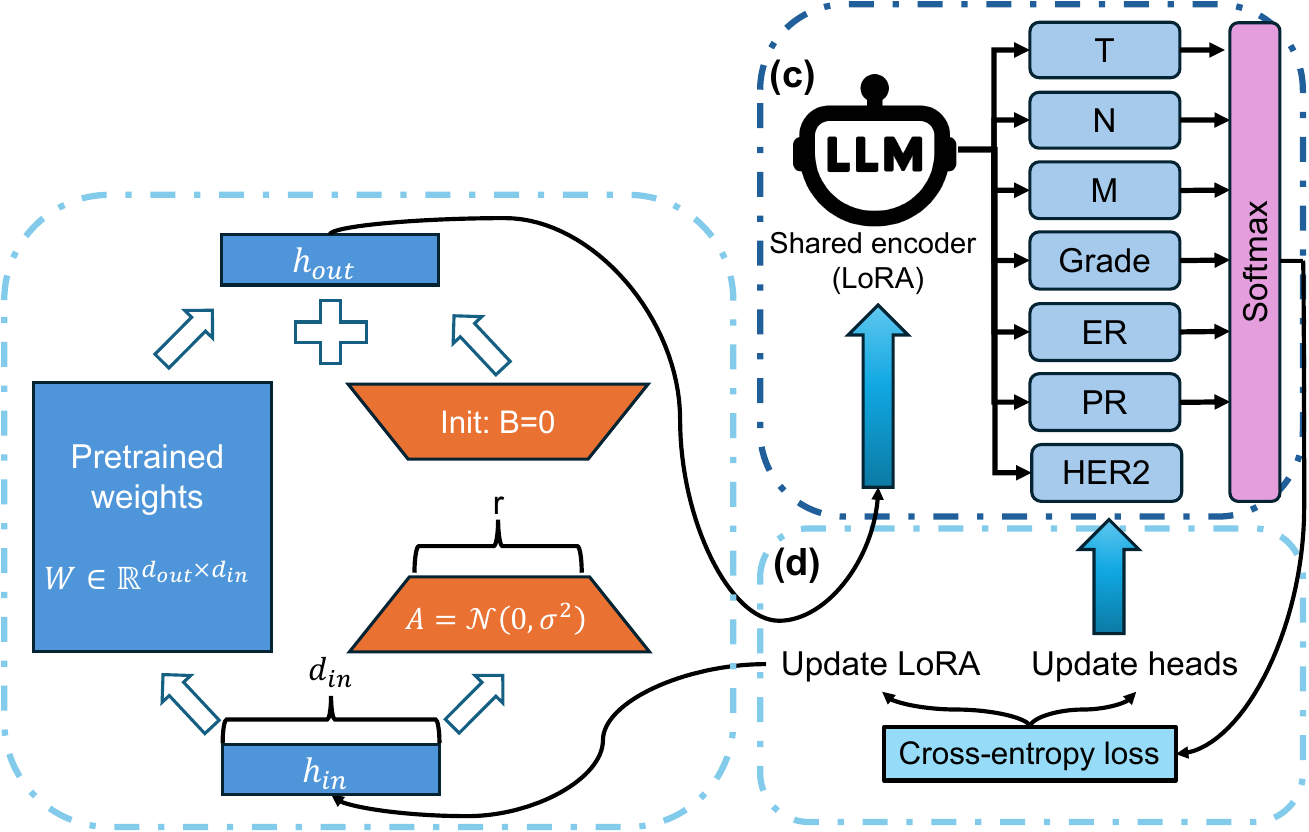} 
            \caption{
            Schematic of the LoRA mechanism. 
            }
            \label{fig:lora}
        \end{figure}
        
        \subsubsection{LoRA Mechanism}
        \label{sec:lora}
        To adapt the general-purpose Llama-3 backbone to the specialized syntax of pathology reports, we employ LoRA \cite{hu2022lora} to avoid the prohibitive computational cost of full-parameter fine-tuning. LoRA is motivated by a key hypothesis: task- and domain-specific adaptations from a strong pre-trained model lie in a low-dimensional subspace of the full parameter space. As illustrated in Fig.~\ref{fig:lora}, to modify the self-attention mechanism of the Llama-3 backbone, we freeze the pre-trained weights and inject trainable low-rank matrices into four projection layers: query ($\boldsymbol{W}_q$), key ($\boldsymbol{W}_k$), value ($\boldsymbol{W}_v$), and output ($\boldsymbol{W}_o$). Formally, let $\boldsymbol{W}_0 \in \mathbb{R}^{d_{out} \times d_{in}}$ represent one of these frozen projection matrices. Here, $d_{in}$ corresponds to the model hidden dimension $d_{model}$ ($d_{model}=4,096$ for Llama-3-8B-Instruct architecture) and $d_{out}$ denotes the projection dimension associated with the corresponding attention head. Instead of optimizing $\boldsymbol{W}_0$ directly, we model the adapted weight matrix as
        \begin{equation}
            \boldsymbol{W}= \boldsymbol{W}_0 + \Delta \boldsymbol{W} 
        \end{equation}
        where the update $\Delta \boldsymbol{W}$ is constrained to have rank at most $r \ll \min(d_{\text{in}}, d_{\text{out}})$. This constraint is enforced structurally by factorizing the update as
        \begin{equation}
            \Delta \boldsymbol{W} = \frac{\alpha}{r}\boldsymbol{B}\boldsymbol{A}
        \end{equation}
        where $\boldsymbol{A}\in \mathbb{R}^{r\times d_{in}}$ and $\boldsymbol{B}\in \mathbb{R}^{d_{out}\times r}$ are the trainable, low-rank adapter matrices. The scaling factor $\frac{\alpha}{r}$ ensures that the magnitude of the low-rank update is appropriately calibrated relative to the frozen backbone $\boldsymbol{W}_0$. Then, given an input hidden state $\boldsymbol{h}_{in} \in \mathbb{R}^{d_{in}}$, the adapted forward pass becomes 
        \begin{equation}
            \boldsymbol{h}_{out} = \boldsymbol{W}_0 \boldsymbol{h}_{in} + \Delta \boldsymbol{W} \boldsymbol{h}_{in}.
        \end{equation}
        
        To ensure the training trajectory begins from the stable pre-trained solution, we initialize $\boldsymbol{A} \sim \mathcal{N}(0, \sigma^2)$ and $\boldsymbol{B} = \mathbf{0}$. This ensures $\Delta \boldsymbol{W} = \mathbf{0}$ at step $t=0$. Let $\boldsymbol{\Theta}_{LoRA}$ denote the complete collection of these adapter matrices ($\boldsymbol{A}$'s and $\boldsymbol{B}$'s) across all adapted transformer layers. The optimization problem reduces to finding the optimal parameters that minimize the objective:
        \begin{equation}
            \min_{\boldsymbol{\Theta}_{LoRA}} \mathcal{L}(\Phi_{frozen} + \Delta \Phi(\boldsymbol{\Theta}_{LoRA}); \mathcal{D}),
        \end{equation}
        where $\mathcal{L}(\cdot)$ is the task-specific loss defined over the pathology report dataset $\mathcal{D}$. The term $\Phi_{frozen}$ denotes the collection of all frozen backbone parameters, and $\Delta \Phi(\boldsymbol{\Theta}_{LoRA})$ denotes the induced parameter updates from the LoRA adapters. 
        By optimizing this constrained subspace while holding $\Phi_{frozen}$ constant, we enable the model to acquire domain-specific clinical features. Importantly, this approach does not corrupt the general linguistic knowledge encoded in the pre-trained backbone, thereby mitigating the catastrophic forgetting issue. 
        
    \subsubsection{Multi-Head Classification and Optimization}
        We develop an MTL framework to exploit the intrinsic semantic correlations among clinical variables. Although T Stage, N Stage, histologic grade, and receptor status represent distinct staging criteria, they are all derived from a unified description of the same biological entity within the pathology report. By training a shared encoder to predict these variables concurrently, we encourage the model to learn a robust semantic representation. This approach offers two key advantages. First, it allows tasks with limited supervision (e.g., rare receptor subtypes) to leverage features learned from data-abundant tasks. This mitigates the impact of class imbalance and prevents overfitting to surface-level artifacts \cite{ruder2017overview}. Second, from a system perspective, this unified architecture minimizes the deployment footprint. It replaces seven standalone models with a single inference pipeline, ensuring scalability in resource-constrained clinical environments.

        To implement the MTL framework, we connect the LoRA-adapted shared encoder to parallel task-specific classification heads. This architecture is depicted in Fig.~\ref{fig:framework}(c). Each head corresponds to one of the target variables in the set $\mathcal{T} = \{T, N, M, Grade, ER, PR, HER2\}$. Formally, let $\boldsymbol{z}^{(p)} = f(\boldsymbol{x}^{(p)};\boldsymbol{\Theta}_{LoRA}) \in \mathbb{R}^{d_{model}}$ denote the high-dimensional latent representation extracted from the terminal token of the input sequence $\boldsymbol{x}^{(p)}$, where $f(\cdot;\boldsymbol{\Theta}_{LoRA})$ represents the shared LoRA-adapted encoder. Given the causal attention mechanism of the Llama-3 backbone, this terminal state serves as a global semantic summary, having attended to the entire preceding clinical narrative and structured hints. 
        This contextual embedding is then projected into independent linear classification heads to generate probability distributions for each task $t$:
        \begin{equation}
            \hat{\boldsymbol{y}}_t^{(p)} = \text{Softmax}(\boldsymbol{W}_t \boldsymbol{z}^{(p)} + \boldsymbol{b}_t), \quad \forall t \in \mathcal{T}
            \label{eq:softmax}
        \end{equation} 
        Here, $\boldsymbol{W}_t \in \mathbb{R}^{C_t \times d_{model}}$ and $\boldsymbol{b}_t \in \mathbb{R}^{C_t}$ denote the learnable weight matrix and bias vector for task $t$. The parameter $C_t$ represents the number of classes for that specific variable. 
        The entire network is trained end-to-end by minimizing the joint Multi-Task Cross-Entropy Loss:
        \begin{equation}
            \mathcal{L}(\boldsymbol{\Theta}; \mathcal{D}) = - \sum_{p=1}^{S} \sum_{t \in \mathcal{T}} \sum_{c=1}^{C_t} \omega_{t,c} \cdot \mathbb{I}(y_{t}^{(p)} = c) \log(\hat{y}_{t,c}^{(p)}),
            \label{eq:loss}
        \end{equation}     
       where $\boldsymbol{\Theta} = \boldsymbol{\Theta}_{LoRA} \cup \{\boldsymbol{W}_t, \boldsymbol{b}_t\}_{t \in \mathcal{T}}$ denotes the complete set of trainable parameters, excluding the frozen backbone, and parameter $\omega_{t,c}$ is introduced to address the class imbalance issue. 

      While several variables exhibit distributional skew (e.g., N Stage, Grade), preliminary experiments indicated that the model could robustly learn these categories without intervention. This robustness is likely attributable to the high semantic distinctiveness of their reporting formats, where staging parameters are frequently demarcated by explicit standardized headers (e.g., \texttt{Regional Lymph Nodes}) or rigid alphanumeric codes (e.g., ``pN0''). 
       However, the HER2 task presents unique challenges: the combination of extreme class imbalance and subtle linguistic cues (e.g., distinguishing "Equivocal" from "Negative") caused the model to overfit the majority class. To mitigate this specific failure mode, we calculated $\omega_{HER2,c}$ using the balanced inverse-frequency formula:
       \begin{equation}
           \omega_{HER2,c} = \frac{S_{valid}}{C_{HER2} \cdot s_{c}}
       \end{equation}
       where $S_{valid}$ is the total number of valid training samples for the task (excluding missing values) and $s_{c}$ is the frequency count of class $c$. This ensures misclassifications of the minority positive class incur a proportionally higher penalty, preventing the model from converging to a trivial majority-baseline. For all other tasks $t \in \mathcal{T} \setminus \{HER2\}$, we assign a uniform weight of $\omega_{t,c}=1$. 

\section{Materials and Experimental Design}
\label{sec:materials}

    \subsection{Clinical Dataset and Cohort Selection}
        We initialize our pipeline with 285,033 pathology reports collected from the Research Enterprise Datawarehouse at University of Tennessee Health Science Center, spanning from 2020 to 2023. This archive is filtered for reports containing structured synoptic variables, especially TNM Stages, histologic grade, and biomarkers. While primarily focused on breast oncology, the cohort intentionally incorporates a diverse range of other malignancies, including lung, colorectal, and prostate cancers to facilitate inductive knowledge transfer across solid tumors sharing AJCC staging logic. 

        To resolve data fragmentation, a patient-level consolidation algorithm is designed to propagate biomarkers across temporal records and concurrent addenda. Following consolidation and pruning of non-informative reports, the final cohort comprised $S=10,677$ unique samples containing at least one clinical staging variable. Ground truth labels are generated using the validated symbolic engine described in Supplementary Appendix I. Based on a manual audit of 7,481 cases (70.1\%), this dataset is designated a High-Fidelity Ground Truth Composite. Data were partitioned using a stratified 80/10/10 strategy for training, validation, and testing. Detailed clinical definitions and long-tail class distributions are provided in Supplementary Appendix IV.

    \subsection{Experimental Design and Comparative Baselines}
    \label{sec:experimental_design}
        The Multi-Task LLM is developed using the Llama-3-8B-Instruct backbone and fine-tuned via 4-bit QLoRA to maximize computational efficiency. The complete configuration of hyperparameters and hardware specifications are documented in Supplementary Appendix II.

\begin{figure}[t]
            \centering
            \includegraphics[width=\columnwidth]{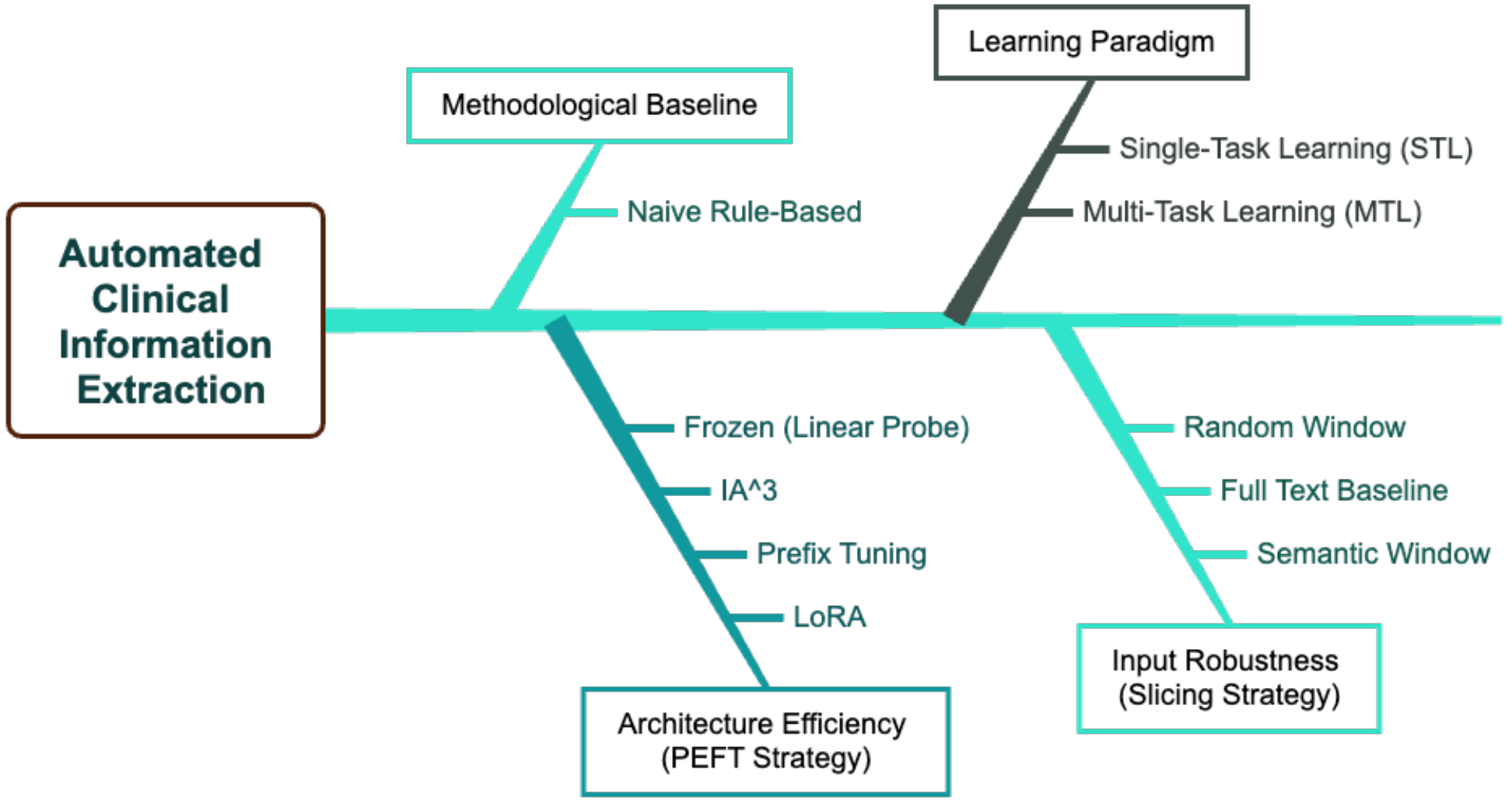}
            \caption{Fishbone diagram of the experimental design and ablation framework. 
            }
            \label{fig:experimental_design}
        \end{figure}

        To evaluate our proposed framework, we implement a systematic experimental design targeting four dimensions: methodological baseline, architectural efficiency, learning paradigm, and input robustness (Fig. \ref{fig:experimental_design}). Methodological Baseline establishes legacy clinical extraction performance via a Naive Rule-Based model applying deterministic regex and proximity-based keyword matching to unsegmented reports. For TNM staging, the algorithm captures standard clinical patterns (e.g., ``pT'' followed by an integer), while for biomarkers, it employs a window-based search ($\pm 75$ characters). This method lacks the context-aware disambiguation utilized in our proposed system, serving as a measure of raw task complexity.
        
        Architectural Efficiency benchmarks our LoRA-adapted Multi-Task LLM (targeting $\boldsymbol{W}_q, \boldsymbol{W}_k, \boldsymbol{W}_v, \boldsymbol{W}_o$) against alternative neural adaptation strategies. We implemented three comparative methods using the Llama-3-8B-Instruct backbone: (1) Frozen (Linear Probing), updating only the classification heads to quantify pre-trained semantic features; (2) (IA)$^3$ \cite{liu2022few}, which rescales inner activations via element-wise multiplication ($\approx 0.01\%$ of parameters). (3) Prefix Tuning \cite{li2021prefix}, which prepends $L_{pre}=32$ learnable virtual tokens to every transformer layer. All neural baselines utilize identical window-slicing inputs and are constrained to a fixed training budget of 3 epochs with a $1 \times 10^{-4}$ learning rate.

        The Learning Paradigm compares our proposed MTL framework with Single-Task Learning (STL), where independent adapters are trained for each variable. This comparison quantifies how a shared semantic encoder facilitates inductive transfer to mitigate extreme class imbalance. Input Robustness further evaluates our Semantic Window Slicing strategy ($\pm 5$ lines surrounding relevant headers) against two baseline conditions: (1) a Full-Text baseline, truncated only by the model's 512-token limit, and (2) a Random Window baseline of equivalent token length. By retaining the \texttt{Hints} block across all conditions, we isolate the windowing mechanism to investigate trade-offs between input density, computational latency, and  semantic completeness. Performance was measured across predictive accuracy, Macro F1, AUROC, training time, and peak GPU memory utilization.

        \begin{table}[!t] 
            \centering
            \caption{Performance comparison on the held-out test set ($S_{test}=1,068$).}
            \label{tab:main_results}
            \scriptsize
            \setlength{\tabcolsep}{0pt} 
            \begin{tabular*}{\columnwidth}{@{\extracolsep{\fill}} l ccc ccc @{}} 
            \toprule
            \multirow{2}{*}{\textbf{Clinical Task}} & \multicolumn{3}{c}{\textbf{Naive Baseline}} & \multicolumn{3}{c}{\textbf{Multi-Task LLM}} \\ 
            \cmidrule(lr){2-4} \cmidrule(l){5-7} 
             & \textbf{Accuracy} & \textbf{F1} & \textbf{AUC} & \textbf{Accuracy} & \textbf{F1} & \textbf{AUC} \\ \midrule
            T Stage & 0.627 & 0.447 & 0.747 & \textbf{0.994} & \textbf{0.993} & \textbf{1.000} \\
            N Stage & 0.911 & 0.739 & 0.933 & \textbf{1.000} & \textbf{1.000} & \textbf{1.000} \\
            Histologic Grade & 0.543 & 0.498 & 0.752 & \textbf{0.969} & \textbf{0.969} & \textbf{0.995} \\
            ER Status & 0.919 & 0.618 & 0.950 & \textbf{0.977} & \textbf{0.961} & \textbf{0.986} \\
            PR Status & 0.937 & 0.635 & 0.964 & \textbf{0.962} & \textbf{0.957} & \textbf{0.990} \\
            HER2 Status & 0.558 & 0.387 & 0.659 & \textbf{0.985} & \textbf{0.967} & \textbf{0.989} \\ \midrule
            \textbf{Overall Average} & 0.749 & 0.554 & 0.834 & \textbf{0.981} & \textbf{0.976} & \textbf{0.996} \\ \bottomrule
            \end{tabular*}
        \end{table}

\section{Results and Discussion}
\label{sec:results}

    \subsection{Performance Comparison with Rule-Based Baselines}
        We benchmark the performance of our Multi-Task LLM against the Naive Rule-Based Baseline on the held-out test set. Table~\ref{tab:main_results} summarizes the performance across the extraction tasks.  Because the clinical cohort utilized in this study consists exclusively of non-metastatic (M0) cases, the M Stage task was excluded from the results tables. According to Table~\ref{tab:main_results}, while the Naive Baseline achieved competitive accuracy for standardized variables like N Stage (0.911) and PR status (0.937), these high accuracy scores overestimate the model's reliability by concealing its failure to distinguish minority classes. For instance, the Macro F1 score of the baseline for PR status is only 0.635, indicating a severe bias toward the majority class. In contrast, our Multi-Task LLM maintains high Macro F1 scores across all tasks, demonstrating robustness to the inherent class imbalances. The performance disparity is most pronounced in tasks requiring contextual disambiguation, specifically T Stage and HER2. For example, the baseline achieves an accuracy of only 0.627 for T Stage, a substantial gap compared to the Multi-Task LLM (0.994). Error analysis reveals that the Naive Baseline frequently captures historical T-stages mentioned in the \texttt{Clinical History} section (e.g., extracting a prior ``T1'' diagnosis) rather than current findings in the \texttt{Final Diagnosis}. Our Multi-Task LLM effectively suppresses these historical mentions, correctly identifying current diagnostic findings in 99.4\% of cases. 

        Similarly, tasks involving complex syntax, such as histologic grade and HER2 status, prove challenging for the rigid baseline. For histologic grade, the baseline conflates the primary tumor grade with unrelated numeric descriptors (e.g., ``Nuclear Grade''), resulting in an accuracy of only 0.543. For HER2, it fails to parse non-standard formats such as ``Equivocal (2+)'', leading to a Macro F1 score of 0.387, which approaches the performance of a trivial majority-class baseline. The Multi-Task LLM generalizes effectively across these heterogeneous formats, achieving a HER2 accuracy of 0.985 and a Macro F1 of 0.967. Even in tasks where the baseline is competitive (e.g., ER status), our Multi-Task LLM provides consistent improvements by capturing the full spectrum of linguistic variation in the reports. 
        
        Moreover, our Multi-Task LLM demonstrates high training efficiency, reaching performance saturation early in the training cycle. Our framework also maintains high sensitivity and specificity on minority classes, effectively resolving the challenges posed by the severe class imbalance issue in biomarker status. Detailed training convergence curves, validation loss plateaus, and the task-specific confusion matrix for HER2 status are provided in Supplementary Appendix III. 
    
    \subsection{Performance Comparison with PEFT Strategies}
        \begin{table}[t]
        \centering
        \caption{Comparison of classification performance and computational resource utilization across PEFT strategies.}
        \label{tab:peft_comparison}
        \scriptsize
        \setlength{\tabcolsep}{0.5pt}
        \resizebox{\columnwidth}{!}{%
        \begin{tabular*}{\columnwidth}{@{\extracolsep{\fill}} l ccc ccc @{}}
        \toprule
        \textbf{Method} & \shortstack{\textbf{Trainable}\\\textbf{Params}} & \shortstack{\textbf{Time}\\\textbf{(s)}} & \shortstack{\textbf{Memory}\\\textbf{(MB)}} & \textbf{Accuracy} & \textbf{Macro F1} & \textbf{AUROC} \\ \midrule
        Frozen (Head) & 81,940 & 5,539 & 5,433 & 0.664 & 0.490 & 0.762 \\
        (IA)$^3$ & 606,228 & 9,198 & 12,553 & 0.673 & 0.512 & 0.756 \\
        Prefix Tuning & 213,012 & 8,517 & 10,716 & 0.978 & 0.972 & 0.994 \\
        \textbf{LoRA (Ours)} & \textbf{13,713,428} & \textbf{3,434} & \textbf{14,211} & \textbf{0.981} & \textbf{0.976} & \textbf{0.996} \\ \bottomrule
        \end{tabular*}
        }
        \end{table}
        Table \ref{tab:peft_comparison} presents a comparative analysis between our LoRA strategy and the alternative adaptation methods. The $\text{IA}^3$ method failed to substantially outperform the Frozen baseline (Accuracy 0.673 vs. 0.664). This suggests that modifying activation magnitudes alone lacks the representational capacity to bridge the semantic gap between general-domain pre-training and the highly specialized, hierarchical schema of oncology staging. 

        In contrast, both Prefix Tuning and LoRA achieved high performance, with accuracies of 0.978 and 0.981, respectively. However, the operational metrics highlight a distinct trade-off between memory footprint and training throughput. Prefix Tuning proved more memory-efficient (10.7 GB vs. 14.2 GB peak VRAM) but suffered from severe computational latency, requiring 8,517 seconds to train, nearly $2.5\times$ slower than LoRA. This latency is likely attributable to the implementation overhead of virtual tokens, which extend the effective sequence length and can disrupt the efficacy of hardware-optimized attention kernels (e.g., Flash Attention) by altering the contiguous memory layout required for tiling operations. While the per-step computational cost of LoRA is marginally higher due to the larger parameter count ($13.7$ million), it achieves optimal convergence in a single epoch (3,434 seconds). In contrast, alternative methods required a full three-epoch schedule to reach peak performance, resulting in a significantly longer total training duration. Although LoRA incurs a higher peak VRAM cost due to the storage of adapter states, this footprint fits comfortably within modern enterprise GPU budgets (e.g., NVIDIA A100 or L40S). 

\subsection{Impact of Multi-Task Learning on Efficiency and Inductive Transfer}
    \label{sec:mtl_ablation}
        \begin{table*}[t]
        \centering
        \caption{Comparative analysis of efficiency, aggregate performance, and task-specific Macro F1 scores}
        \label{tab:mtl_vs_stl_combined}
        \footnotesize 
        \setlength{\tabcolsep}{0pt} 
        \begin{tabular*}{\textwidth}{@{\extracolsep{\fill}} lcccccccccccc}
        \toprule
         & \multicolumn{3}{c}{\textbf{Resource Efficiency}} & \multicolumn{3}{c}{\textbf{Aggregate Performance}} & \multicolumn{6}{c}{\textbf{Task-specific Macro F1}} \\ 
        \cmidrule(lr){2-4} \cmidrule(lr){5-7} \cmidrule(l){8-13}
        \textbf{Architecture} & \textbf{Models} & \textbf{Params} & \textbf{Time} & \textbf{Accuracy} & \textbf{Macro F1} & \textbf{AUROC} & \textbf{T} & \textbf{N} & \textbf{Grade} & \textbf{ER} & \textbf{PR} & \textbf{HER2} \\ \midrule
        STL Baseline & 6 & 82.2 M & 3,473 s & 0.928 & 0.914 & 0.946 & 0.991 & \textbf{1.000} & \textbf{0.970} & 0.919 & 0.900 & 0.309 \\
        \textbf{MTL (Ours)} & \textbf{1} & \textbf{13.7 M} & \textbf{3,434 s} & \textbf{0.981} & \textbf{0.976} & \textbf{0.996} & \textbf{0.993} & \textbf{1.000} & 0.969 & \textbf{0.961} & \textbf{0.957} & \textbf{0.967} \\ 
        \bottomrule
        \end{tabular*}
        \end{table*}
        To validate the structural advantages of our unified architecture, we further benchmark the MTL framework against a STL baseline. In the STL configuration, we train six independent LoRA adapters, one for each clinical variable analyzed in the results, maintaining identical hyperparameters to the joint model. Table \ref{tab:mtl_vs_stl_combined} summarizes the comparative resource utilization,  aggregate predictive power, and task-specific performance across both architectures. Our MTL approach demonstrates superior efficiency, reducing the deployment footprint by a factor of six ($13.7$ million vs. $82.2$ million trainable parameters). This consolidation ensures scalability for resource-constrained clinical environments while maintaining comparable training latency. Additionally, the unified model achieves superior aggregate metrics, raising the mean Accuracy from 0.928 to 0.981 and Macro F1 from 0.914 to 0.976. 

        The granular breakdown of task-specific performance reveals that while Single-Task models performs competently on data-rich variables like N Stage and T Stage, the STL baseline exhibits a performance collapse on HER2 status, yielding a Macro F1 score of only 0.309. This task is characterized by high class imbalance (as detailed in Supplementary Material II) and limited training support for positive cases, exhibiting a severe 6.4:1 imbalance ratio. The results indicate that the isolated HER2 model, lacking sufficient supervision signals, fails to generalize. In stark contrast, our Multi-Task framework achieves a Macro F1 of 0.967 on HER2. This confirms that the joint training objective facilitates effective inductive knowledge transfer, enabling the data-scarce HER2 task (with 1,597 training samples) to leverage high-level semantic features learned from data-abundant tasks such as T Stage and N Stage, with 6,240 and 3,947 training samples, respectively. 

    \subsection{Input Robustness and Semantic Slicing Ablation}
        Table \ref{tab:ablation} presents the impact of the input processing strategies described in Research Methodology section. Note that the Llama-3 architecture demonstrated remarkable robustness to unstructured noise introduced by the unpruned narratives containing extensive administrative metadata, with the Full Text configuration achieving a Macro F1 of 0.977, which is statistically on par with our Window Slicing approach. Despite this parity, the Window Slicing strategy is the preferred deployment configuration because it mitigates the quadratic computational overhead of self-attention ($\mathcal{O}(L^2)$) by significantly reducing the average input sequence length, which directly lowers inference latency. Moreover, the validity of our slicing strategy is confirmed by comparing with the Random Window baseline with Accuracy of 0.854 and Macro F1 of 0.790. This is due to the fact that relevant context might be removed in Random Window slicing. This sharp deterioration confirms that the model's high performance relies on attending to specific, semantically relevant diagnostic segments, whether identified implicitly via global attention or extracted explicitly by our slicer. 

        \begin{table}[t]
        \centering
        \caption{Ablation study on Input Robustness.}
        \label{tab:ablation}

        \setlength{\tabcolsep}{3pt} 
        \footnotesize 
        \begin{tabularx}{\columnwidth}{Xccc}
        \toprule
        \textbf{Configuration} & \textbf{Accuracy} & \textbf{Macro F1} & \textbf{AUROC} \\ \midrule
        Full Text (No Slicing) & \textbf{0.982} & \textbf{0.977} & 0.995 \\
        Random Window Baseline & 0.854 & 0.790 & 0.921 \\
        \textbf{\mbox{Window Slicing ($\pm 5$ lines)}} & 0.981 & 0.976 & \textbf{0.996} \\
        \bottomrule
        \end{tabularx}
        \end{table}
    
\section{Conclusion}
This paper proposes a novel Multi-Task LLM framework for automated breast cancer staging and biomarker extraction from pathology narratives. By repurposing the Llama-3-8B-Instruct model as a discriminative encoder and leveraging LoRA, the framework effectively addresses the prohibitive computational costs of full-parameter fine-tuning and brittle rule-based systems. Specifically, the parallel classification architecture ensures consistent schema adherence by replacing the auto-regressive decoding with classification heads. Concurrently, the MTL architecture facilitates inductive knowledge transfer, effectively mitigating severe class imbalance challenges for data-scarce variables. Experimental results demonstrate that the framework achieves a Macro F1 score of 0.976, significantly outperforming legacy rule-based baselines and single-task LLM counterparts. Our framework provides healthcare systems with a scalable, high-fidelity tool for modernizing cancer registries and accelerating data-driven oncology research, with a methodology broadly applicable to other clinical domains that require robust, evidence-based information extraction.

 \section*{Data availability Statement}
 This study was approved by the Institutional Review Board (IRB) at University of Tennessee Knoxville, approval ID \#5156. The study was conducted in accordance with relevant institutional guidelines and regulations. 
 The clinical pathology reports were sourced from the Research Enterprise Datawarehouse at the University of Tennessee Health Science Center. These data contain protected health information and are not publicly available due to institutional privacy policies and HIPAA regulations. De-identified data and the extraction logic may be made available to qualified researchers upon reasonable request to the corresponding author, subject to IRB approval and the execution of a Data Use Agreement.




\section{Acknowledgment}
 This research has been supported by a seed grant from the Human Health and Wellness Gateway at the University of Tennessee, Knoxville.
 We acknowledge with deep gratitude Dr. John Bell, former Director of the Cancer Institute at the University of Tennessee Medical Center, whose mentorship, collaboration, and original initiation of this work were central to its development. His intellectual leadership and generosity continue to inspire us. This paper is dedicated to his memory.

\bibliographystyle{IEEEtran}
\bibliography{references}

\clearpage

\onecolumn 
\pagenumbering{arabic} 
\renewcommand{\thepage}{S\arabic{page}} 
\renewcommand{\thesection}{S\arabic{section}} 
\renewcommand{\thefigure}{S\arabic{figure}} 
\renewcommand{\thetable}{S\arabic{table}} 
\setcounter{section}{0}
\setcounter{figure}{0}
\setcounter{table}{0}

\begin{center}
    \LARGE \textbf{Supplementary Material}
\end{center}

\vspace{0.3cm}
\noindent This supplementary document provides additional details that support the main manuscript titled:  \textit{``Multi-Task LLM with LoRA Fine-Tuning for Automated Cancer Staging and Biomarker Extraction''}

\section{Data Preprocessing and Ground Truth Construction}

The initial phase converts raw, heterogeneous pathology documents into anonymized, machine-readable text. As depicted in Figure 1(a) of the main manuscript, we prioritize native digital text extraction to handle the variability of clinical formats while preserving original character encodings. While the majority of records are digital-born and parsed via \texttt{PyPDF2}, the system includes a contingency module to handle flattened records or documents with unmapped font encodings. This fallback ensures dataset continuity by utilizing an image-based extraction path, powered by \texttt{EasyOCR}, for the documents that fail standard parsing. This ensures that the clinical narrative is recovered accurately before proceeding to the de-identification and ground-truth generation phases \cite{pypdf2, easyocr}. Following extraction, the raw text undergoes strict de-identification using automated pattern matching to redact Protected Health Information (PHI). This includes direct identifiers such as names, medical record numbers, and dates of birth, ensuring compliance with privacy standards (HIPAA). The input is strictly anonymized prior to analysis.

Finally, we address ambiguity in multi-specimen reports, such as those describing concurrent bilateral breast procedures. We implement a regex-driven segmentation module that scans the text stream for a prioritized set of anchor headers. These include \texttt{SYNOPTIC SUMMARY (LEFT/RIGHT BREAST)} and \texttt{INVASIVE DUCTAL/LOBULAR CARCINOMA}. By resolving the start indices of these headers, the system partitions the document into discrete diagnostic blocks. This spatial slicing ensures that the subsequent input text provided to the model contains only the relevant clinical context for a specific tumor site, preventing feature cross-contamination from adjacent diagnoses.

\subsection{Expert-Verified Ground Truth Construction}
\begin{figure*}[t]
    \centering            
    \includegraphics[width=0.7\textwidth]{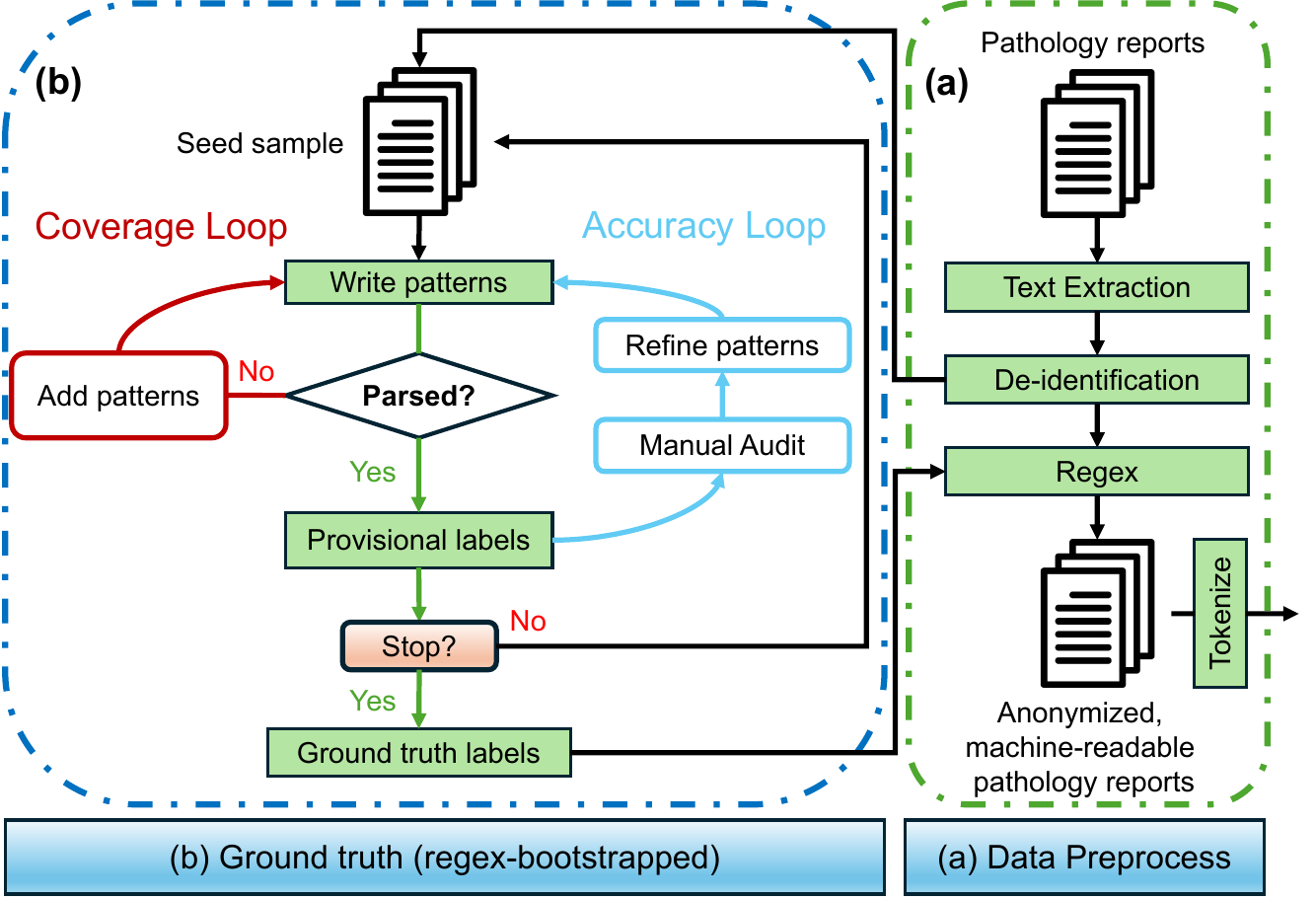}
    \caption{
            Detailed workflow of the ground truth generation process. (a) \textbf{Data Preprocessing:} The pipeline begins with native PDF text extraction and de-identification to produce machine-readable text. (b) \textbf{Human-Verified Bootstrapping:} The core iterative loop initialized with a seed sample. The process splits into two feedback cycles: a \textbf{Coverage Loop} (red arrows) for unparsed reports, and an \textbf{Accuracy Loop} (blue arrows) where provisional regex outputs are rigorously validated via manual expert audit against the source documents. The cycle repeats until the logic stabilizes, yielding no further rule refinements on the manually verified cohort.
            }
    \label{fig:groundtruth}
\end{figure*}

To establish high-quality ground truth labels $\boldsymbol{y}^{(p)}$ at scale, we developed an iterative regex-based labeling engine based on the semi-automated human-in-the-loop strategy illustrated in Figure \ref{fig:groundtruth}. We first analyzed a seed sample to construct a hierarchical library of regex patterns. Unlike simple keyword matching, this engine employs context-aware priority logic to resolve conflicts. For instance, to distinguish between historical data and the current diagnosis, the logic prioritizes values found under explicit section headers (e.g., \texttt{FINAL DIAGNOSIS}) over conflicting mentions in summary lines or clinical histories. Furthermore, specific negative assertions (e.g., ``No evidence of malignancy'') are prioritized to override keyword matches for tumor types found elsewhere in the document. A canonicalization module subsequently maps heterogeneous clinical descriptions into standardized class labels. Examples include mapping Roman numerals (``Grade III''), fractional scores (``3/3''), or qualitative terms (``Poorly Differentiated'') into uniform G3 classification.

As depicted in Figure \ref{fig:groundtruth} (b) of the main manuscript, the workflow integrates two distinct quality assurance cycles. The Coverage Loop (red arrows) identifies unparsed reports to capture unhandled formats and edge cases. Simultaneously, the Accuracy Loop (blue arrows) involves a rigorous Manual Audit phase where provisional regex outputs are cross-verified against the original source documents by domain experts. This iterative refinement process ensures that the expert-level logic, such as negation handling and assay prioritization, is accurately captured within the regex engine before generating the final ground truth labels $\boldsymbol{y}^{(p)}$. To validate the integrity of this iterative process, we performed a comprehensive manual audit of 7,481 pathology reports (70.1\% of the cohort). During this phase, we refined the regex logic under clinical expert guidance to maximize sensitivity (minimizing extraction misses) and resolve complex linguistic ambiguities, such as distinguishing addendum updates from primary diagnoses. This dual-loop process was repeated until the engine achieved performance saturation, demonstrating near-perfect alignment with human clinical judgment across the verified cohort.

We utilize these verified regex-derived labels as high-fidelity supervision targets, establishing an automated regex-supervised training strategy. By fine-tuning the Multi-Task LLM on this dataset, the model internalizes complex clinical rules within its learned parameters, effectively approximating the decision logic of the iterative regex engine. Crucially, while the regex engine is used as a deterministic labeling oracle during the training phase, the final extraction and inference are performed solely by the neural model. This approach demonstrates that a neural network architecture can successfully capture expert diagnostic logic without requiring the deployment of a cumbersome rule-based engine in the final inference pipeline.

\subsection{Structured Input Representation}
To optimize computational efficiency and input density, we applied a window-slicing strategy extracting a context window of $\pm K$ lines (where $K=5$) around prioritized anchor headers (e.g., \texttt{FINAL DIAGNOSIS}). This constructs a dense input representation $\boldsymbol{x}^{(p)}$ that reduces token usage while retaining high-clinical context. We utilize a structured four-block schema to serve as a bridge between unstructured clinical narratives and the model's extraction heads: (1) \textit{Field}, identifying the task; (2) \textit{Classes}, derived from American Joint Committee on Cancer (AJCC) standards to constrain the output space; (3) \textit{Report}, containing the sliced narrative; and (4) \textit{Hints}, providing raw symbolic cues.

Complementing this structural constraint, we generate hints via a symbolic rule-based engine that localizes task-relevant headers through deterministic pattern matching. For standard fields like T/N Stage, the hint captures the raw substring immediately following the section header (e.g., extracting ``pT1c'' after \texttt{Primary Tumor:}). For complex biomarkers like HER2, the module identifies and concatenates fragmented text segments from multiple heterogeneous diagnostic assays, such as Immunohistochemistry (IHC) intensity and Fluorescence In Situ Hybridization (FISH) amplification status, into a single raw string (e.g., \texttt{HER2\_hint: IHC 2+, FISH amplified}). Rather than attempting to deterministically reconcile these heterogeneous signals using handcrafted rules, the system preserves the original assay-level evidence and presents it directly to the LLM model.

The LLM model must learn to filter, contextualize, and reason over these noisy cues. Because pathology reports are often lengthy and multi-focal, these hints serve as localized semantic anchors that guide the model's self-attention mechanism toward clinically salient regions. This hint format conditions the shared encoder’s hidden representations without imposing rule-based constraints. We pool the resulting contextualized embeddings by extracting the hidden state corresponding to the terminal token of the input sequence. By selecting this terminal state, we ensure the representation captures the global semantic integration across the entire structured prompt.

\section{Implementation Details and Training Configuration}

We implemented the Multi-Task LLM framework using PyTorch and the Hugging Face PEFT library. We initialized the encoder with the official \texttt{meta-llama/Meta-Llama-3-8B-Instruct} checkpoint. To maximize computational efficiency on available hardware, we utilized 4-bit NormalFloat quantization (QLoRA) \cite{dettmers2023qlora}, which significantly reduces memory footprint without degrading predictive performance. All experiments were conducted on a single NVIDIA Ada Lovelace (L40S) GPU (48 GB VRAM) with a fixed random seed (42).

For hyperparameter selection, we adopted standard configurations from established parameter-efficient fine-tuning benchmarks \cite{hu2022lora}. We fixed the rank at $r=16$ and the scaling factor at $\alpha=16$. Low-rank adapters were applied to the self-attention projection matrices ($\boldsymbol{W}_q, \boldsymbol{W}_k, \boldsymbol{W}_v, \boldsymbol{W}_o$). Empirical studies suggest that low-rank adaptations within this range are sufficient to capture domain-specific semantic shifts without the overfitting risks associated with higher-rank updates \cite{hu2022lora, lialin2023scaling}. This targeted configuration resulted in approximately $13.7$ million trainable parameters ($\approx 0.34\%$ of total weights). To ensure robust generalization, we applied a dropout rate of $0.05$ and a weight decay of $0.01$.

The model was trained for a maximum of 3 epochs using the Paged AdamW (8-bit) optimizer with a peak learning rate of $1 \times 10^{-4}$ and a cosine annealing schedule (3\% linear warmup). While the Llama-3 architecture is format-agnostic, we adopted this structured schema as a methodological design choice to align the input with the structural inductive bias of the pre-trained instruction priors \cite{dubey2024llama}. To ensure numerical stability, we utilized bfloat16 (bf16) compute precision. While the model weights were stored in 4-bit NormalFloat (NF4) to minimize VRAM usage, all forward and backward passes were executed in bf16 to leverage its wider dynamic range, which is critical for preventing gradient underflow in large-scale models. We enforced a global maximum sequence length of $L=512$ tokens, truncating longer inputs to fit the context window. To accommodate GPU memory limits, we employed a micro-batch size of 2 with 8 gradient accumulation steps, simulating a stable effective batch size of 16 \cite{dettmers2023qlora}. By aggregating the gradients from eight consecutive forward and backward passes before performing a single weight update, this approach ensures that the optimization trajectory remains robust despite the restricted per-step memory footprint. Given the rapid convergence observed in preliminary trials, we employed an early stopping strategy based on validation loss to prevent overfitting and minimize computational waste.

The experimental design and ablation framework targets four critical dimensions: methodological baseline, architectural efficiency, learning paradigm, and input robustness, as summarized in the fishbone diagram in Figure 3 of the main manuscript.

\section{Results Discussion and Extended Training Analysis}

\begin{figure*}[t]
    \centering
    \includegraphics[width=0.85\textwidth]{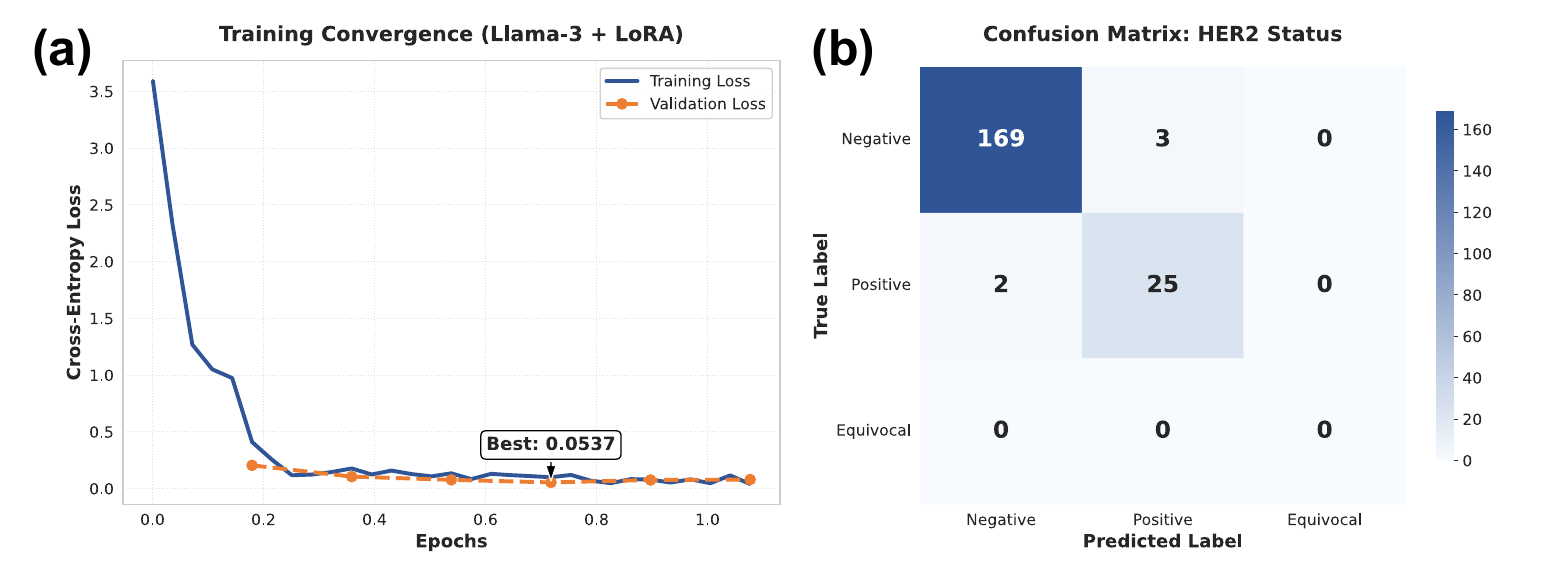}
    \caption{(a) Training Efficiency and (b) HER2 Classification Performance. 
            }
    \label{fig:S3_analysis}
\end{figure*}

To further validate training efficiency, we analyzed the learning trajectory of the Multi-Task LLM (see Figure \ref{fig:S3_analysis} (a)). The model exhibits rapid convergence, reaching a validation loss plateau of 0.0537 within the first epoch. While we maintained a uniform 3-epoch budget for experimental consistency, these results confirm that LoRA-based adaptation is highly efficient for clinical domain-tuning, offering a significant throughput advantage for hospital deployment. This efficiency does not compromise performance; as shown in the HER2 confusion matrix in Figure \ref{fig:S3_analysis} (b), the model correctly identified 92.6\% of positive cases in a highly imbalanced subset. This confirms that the low-rank updates to the self-attention matrices are sufficient to capture complex diagnostic nuances without overfitting. The consolidation of tasks into a single architecture ensures scalability for resource-constrained clinical environments while maintaining high-fidelity extraction across correlated clinical variables.

\section{Clinical Variable Definitions and Class Distribution}
\label{sec:appendix_distribution}
\begin{figure*}[t]
    \centering
    \includegraphics[width=0.85\textwidth]{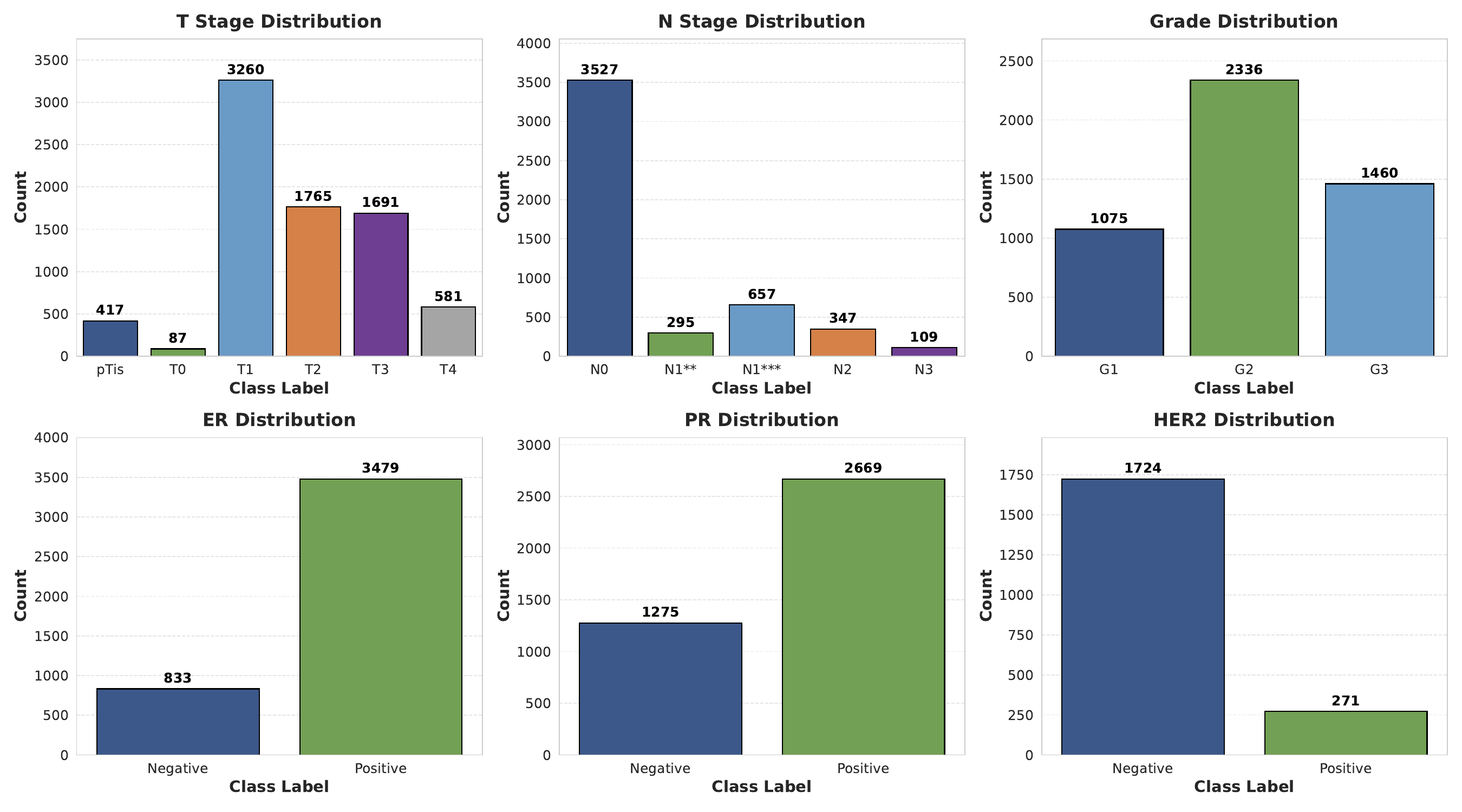}
    \caption{\textbf{Class Distributions across Clinical Tasks.} The histograms illustrate the frequency of ground truth labels for (a) T Stage, (b) N Stage, (c) Histologic Grade, (d) ER, (e) PR, and (f) HER2. 
    }
    \label{fig:class_dist}
\end{figure*}

Table \ref{tab:class_definitions} provides the clinical definitions for the cancer staging variables, derived from the AJCC TNM Staging System (8th Ed.) and National Comprehensive Cancer Network (NCCN) Clinical Practice Guidelines in Oncology (v6.2024) \cite{amin2017ajcc, national2024nccn}. To further characterize the clinical composition of our study cohort ($S=10,677$), we provide a detailed breakdown of label frequencies across the six targeted variables. This distribution, illustrated in Fig.~\ref{fig:class_dist}, reflects the inherent variability of a real-world clinical population treated at a regional oncology center. The histograms reveal a pronounced ``long-tail'' distribution in tumor staging, where earlier stages (``T1'', ``N0'') dominate the cohort. This reflects contemporary early-detection trends in breast cancer screening. Conversely, advanced stages (e.g., ``T4'', ``N3'') and specific biomarker states (``HER2-positive'') represent a small fraction of the total population. These minority classes represent significant challenges for automated extraction. Infrequent clinical findings, such as ``T4'' or ``HER2-positive'' status, provide sparse training data, making it difficult for models to learn their specific linguistic patterns. Furthermore, pathology reports often contain historical staging information, for example, a prior ``T1'' diagnosis mentioned alongside a current ``T3'' finding, that can mislead simpler systems. Successfully resolving these cases requires the model to differentiate between past and present clinical status, a task where rigid, keyword-based approaches typically fail.

\begin{table*}[tb]
    \centering
    \caption{\textbf{Clinical Definitions of Extracted Variables.} Definitions derived from AJCC TNM Staging System (8th Ed.) and NCCN Guidelines (v6.2024). Asterisks (*, **, ***) denote specific prognostic grouping conventions observed in the dataset labels.}
    \label{tab:class_definitions}
    \footnotesize
    \renewcommand{\arraystretch}{1.4}
    \begin{tabularx}{0.9\textwidth}{l l X}
    \toprule
    \textbf{Variable} & \textbf{Class Label} & \textbf{Clinical Definition} \\
    \midrule
    \multirow{6}{*}{T Stage} & pTis & Ductal carcinoma \textit{in situ} (early stage, non-invasive). \\
     & T0 & No evidence of primary tumor. \\
     & T1* & Small tumor ($\le$ 20mm). Includes microinvasive cases ($\le$ 1mm). \\
     & T2 & Medium tumor ($>$ 20mm but $\le$ 50mm). \\
     & T3 & Large tumor ($>$ 50mm). \\
     & T4 & Tumor of any size that has spread directly to the chest wall or skin. \\
    \midrule
    \multirow{5}{*}{N Stage} & N0 & No cancer found in regional lymph nodes. \\
     & N1** & \textbf{Macrometastases} in 1–3 axillary lymph nodes (standard spread). \\
     & N1*** & \textbf{Micrometastases (N1mi)}; small cancer cells ($>$ 0.2mm but $\le$ 2.0mm). \\
     & N2 & Metastases in 4–9 axillary lymph nodes; or nodes are fixed/matted together. \\
     & N3 & Extensive spread ($\ge$ 10 nodes) or spread to collarbone/internal chest nodes. \\
    \midrule
    \multirow{2}{*}{M Stage} & M0 & No clinical or radiographic evidence of distant metastases. \\
     & M1 & Distant metastases detected by clinical, radiographic or histologically means. \\
    \midrule
    \multirow{3}{*}{Grade} & G1 & Low Grade (Well-differentiated); cells look normal and grow slowly. \\
     & G2 & Intermediate Grade (Moderately differentiated); growing faster than normal. \\
     & G3 & High Grade (Poorly differentiated); cells look abnormal and grow aggressively. \\
    \midrule
    \multirow{3}{*}{\shortstack[l]{Biomarkers\\ \footnotesize (ER, PR, HER2)}} 
     & Positive & Cells have receptors that fuel cancer growth (responds to targeted therapy). \\
     & Negative & Cells lack these receptors (requires different treatment approaches). \\
     & Equivocal & Borderline result requiring further testing (e.g., genetic amplification test). \\
    \bottomrule
    \end{tabularx}
\end{table*}

\end{document}